\documentclass[12pt, review]{elsarticle}

\usepackage{amssymb}
\usepackage{diagbox}
\usepackage{makecell, soul}
\usepackage{amsmath,amssymb,amsfonts}
\usepackage{algorithmic}
\graphicspath{{images/}}
\usepackage{textcomp}
\usepackage{xcolor, rotating}
\usepackage{balance}
\usepackage{multirow}
\usepackage{gensymb}
\def\BibTeX{{\rm B\kern-.05em{\sc i\kern-.025em b}\kern-.08em
		T\kern-.1667em\lower.7ex\hbox{E}\kern-.125emX}}

\usepackage{nohyperref}	%pageanchor=false,
\hypersetup{colorlinks, pdfa=true, breaklinks, citecolor=black, urlcolor=black, linkcolor=black}
\usepackage{amssymb,amsmath,epsfig,latexsym, bm}
\usepackage{hyphenat}

\usepackage{mathtools, mdwmath, mdwtab, multirow, amssymb, amsmath, array, ragged2e}

\usepackage{lipsum}
\usepackage{threeparttable, booktabs, url, eqparbox, xspace, setspace, tabularx, subfig}
\makeatletter
\DeclareRobustCommand\onedot{\futurelet\@let@token\@onedot}
\def\@onedot{\ifx\@let@token.\else.\null\fi\xspace}
 
\def\ie{\emph{i.e}\onedot}

\def\etal{\emph{et al}\onedot}
\makeatother

\journal{Neurocomputing}

\begin{document}

\begin{frontmatter}

%% Title, authors and addresses 

%% use the tnoteref command within \title for footnotes;
%% use the tnotetext command for theassociated footnote;
%% use the fnref command within \author or \address for footnotes;
%% use the fntext command for theassociated footnote;
%% use the corref command within \author for corresponding author footnotes;
%% use the cortext command for theassociated footnote;
%% use the ead command for the email address,
%% and the form \ead[url] for the home page:
\title{BLPnet: A new DNN model and Bengali OCR engine for Automatic License Plate Recognition}
%% \tnotetext[label1]{}
%% \author{Name\corref{cor1}\fnref{label2}}
%% \ead{email address}
%% \ead[url]{home page}
%% \fntext[label2]{}
%% \cortext[cor1]{}
%% \affiliation{organization={},
%%             addressline={},
%%             city={},
%%             postcode={},
%%             state={},
%%             country={}}
%% \fntext[label3]{}

\author{Md. Saif Hassan Onim} % \orcidID{0000-0002-7228-2823}
    \ead{saif@eece.mist.ac.bd}
\author{Hussain Nyeem} %\orcidID{0000-0003-4839-5059}%\fnref{hn}
    \ead{h.nyeem@eece.mist.ac.bd}
\author{Koushik Roy}
    \ead{rkoushikroy2@gmail.com}
\author{Mahmudul Hasan}
    \ead{mahmud108974@gmail.com}
\author{Abtahi Ishmam}
    \ead{abtahiishmam3@gmail.com}
\author{Md. Akiful Hoque Akif} % \fnref{dc}
    \ead{mohammadaxif5717@gmail.com}
\author{Tareque Bashar Ovi}
    \ead{ovitareque@gmail.com}

\affiliation{organization={Department of EECE, Military Institute of Science and Technology (MIST)},%Department and Organization
            addressline={Mirpur Cantonment}, 
            city={Dhaka},
            postcode={1216}, 
            state={},
            country={Bangladesh}}

% \affiliation[inst1]{organization={Org of Ovi},%Department and Organization
%             addressline={Address One}, 
%             city={Dhaka},
%             postcode={1216}, 
%             state={},
%             country={Bangladesh}}

\begin{abstract}
%% Text of abstract
The development of the Automatic  License  Plate  Recognition  (ALPR)  system has received much attention for the English license plate.  However, despite being the sixth largest population around the world,  no significant progress can be tracked in the  Bengali language countries or states for the  ALPR system addressing their more alarming traffic management with inadequate road-safety measures.   This paper reports a  computationally efficient and reasonably accurate  Automatic  License  Plate  Recognition  (ALPR)  system for  Bengali characters with a  new end-to-end  DNN  model that we call Bengali License Plate Network(BLPnet).   The cascaded architecture for detecting vehicle regions prior to vehicle license plate (VLP) in the model is proposed to eliminate false positives resulting in higher detection accuracy of  VLP.  Besides,  a  lower set of trainable parameters is considered for reducing the computational cost making the system faster and more compatible for a real-time application.  With a Computational Neural Network (CNN)based new  Bengali  OCR  engine and word-mapping process,  the model is characters rotation invariant,  and can readily extract,  detect and output the complete license plate number of a  vehicle.   The model feeding  with17 frames per second (fps) on real-time video footage can detect a vehicle with the  Mean  Squared  Error  (MSE)  of  0.0152,  and the mean license plate character recognition accuracy of 95\%.  While compared to the other models,  an improvement of  5\%  and  20\%  were recorded for the  BLPnetover the prominent YOLO-based ALPR model and the Tesseract model for the number-plate detection accuracy and time requirement, respectively.

\end{abstract}

%%Graphical abstract
\begin{graphicalabstract}
\\
\fboxsep=1mm%padding thickness
\fboxrule=4pt%border thickness
\fcolorbox{yellow}{white}{%
\includegraphics[width=0.915\linewidth]{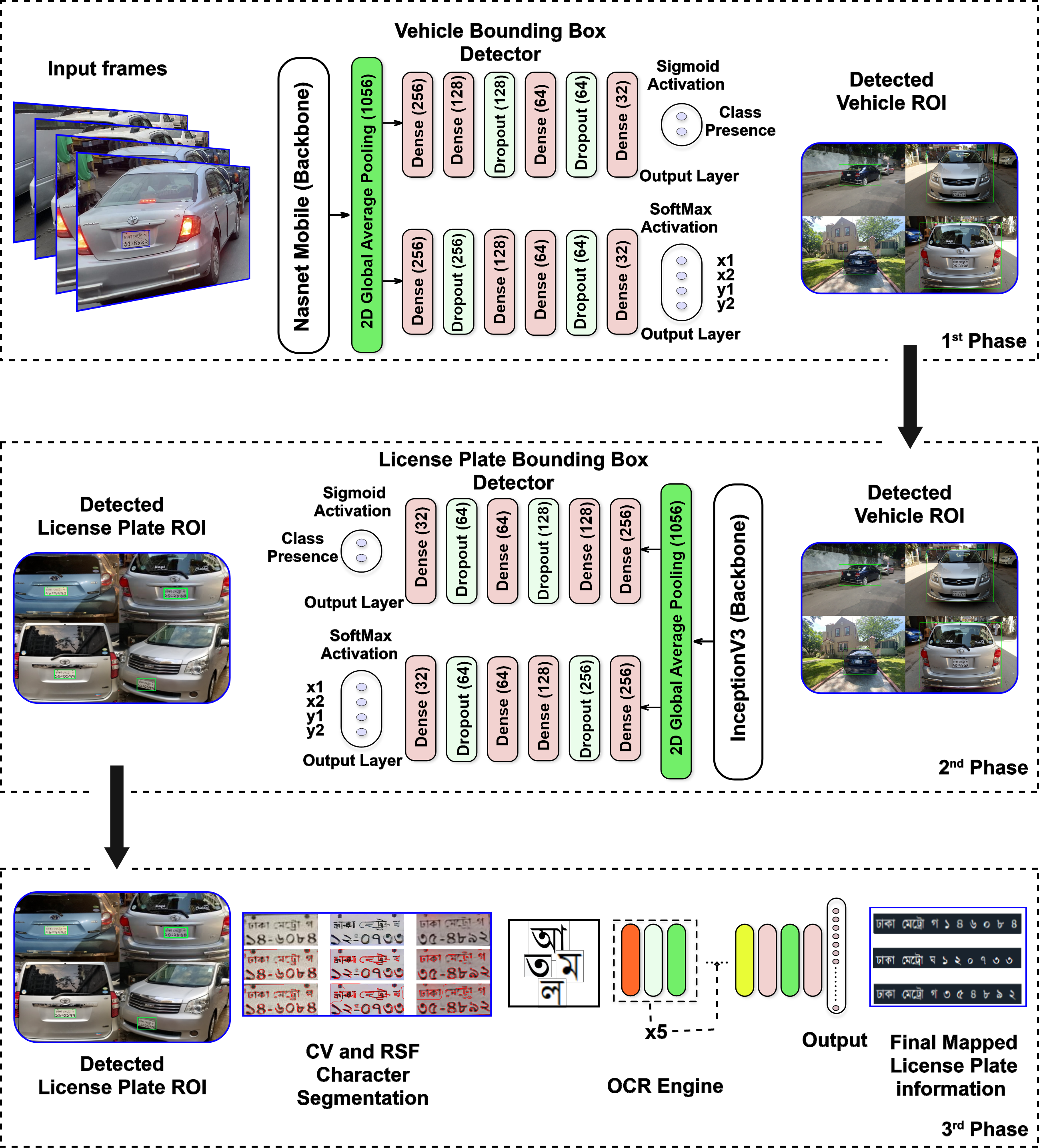}
}
\end{graphicalabstract}

%%Research highlights

\begin{highlights}

    \item We propose an end\hyp to\hyp end Deep Neural Network (DNN) model that we call BLPnet for real\hyp time Automatic License Plate Recognition (ALPR) system.
    
    \item BLPnet is proposed to operate in two separate detection phases to eliminate the false positive detection of number plate.
    
    \item BLPnet also captures a lower set of trainable parameters adequate for the vehicle license plate detection, and thus, its potential for the real\hyp time application is confirmed with its impressive computational efficiency compared to the prominent ALPR systems.
    
    \item A new Convolutional Neural Network (CNN) based OCR engine is developed for rotation invariant Bengali license plate's character recognition  and information retrieval with notably higher accuracy. 
\end{highlights}

\begin{keyword}
%% keywords here, in the form: keyword \sep keyword
ALPR \sep CNN\sep INCEPTION-V3\sep License plate detection\sep NASNet Mobile\sep OCR
% %% PACS codes here, in the form: \PACS code \sep code
% \PACS 0000 \sep 1111
% %% MSC codes here, in the form: \MSC code \sep code
% %% or \MSC[2008] code \sep code (2000 is the default)
% \MSC 0000 \sep 1111
\end{keyword}

\end{frontmatter}

%% \linenumbers
%% main text

%%%%%%%%%%%%%%%%%%%%%%%%%%%%%%%%%%%%%%%%%%%%%%%%%%%%%%%%%
\section{Introduction}
\label{sec-intro}
Automatic License Plate Recognition (ALPR) systems have received much attention to modern transportation services mainly for  automatic management of traffic, parking, toll\hyp station, and other road operations including surveillance and recognition of potential threats~\cite{du2012automatic}.
% field of transportation providing a wide range of possible applications such as tracking of etc. 
An ALPR system consists of three main phases: \textit{frame\hyp selection}, \textit{character\hyp segmentation}, and\textit{ optical character recognition} (OCR).
%~\cite{chen2019automatic}
% \hl{Notes on the Bengali language and its uses in vehicle, overall significance for the Bengali VLP Recognition}
The first phase verifies the existence of any Vehicle License Plate (VLP) in the input frame by extracting any possible character's features from the frame. The second phase separates the characters from the background followed by their recognition in the last phase.
%
% In spite of the existing advanced computer vision technologies and artificial intelligence algorithms, 
% there is still sufficient space for a standard approach, which can tackle even the most difficult issues. 
% Picture selection, object identification, segmentation, and optical character recognition are the four stages in ALPR system\cite{chen2019automatic}.
% 
% The application of ALPR techniques in the
Such a system has more potential for the developing countries or states
requiring higher road\hyp safety measures and better traffic management.\looseness -1
% , which  has remained a key challenge.

One such potential group of the developing regions is the Bengali speaking  countries and states that still requires a promising ALPR system for the Bengali VLP application.
Unlike English characters, Bengali has more complex features leaving its accurate recognition from a VLP more challenging. Despite being the six largest population  around the world~\cite{ ethnologue2021}, no significant progress can be tracked in the Bengali language countries or states for the ALPR system addressing their inadequate road\hyp safety measures and poor traffic management. Besides, the performance of the prominent ALPR models developed for the English VLP is also unknown for the Bengali VLP recognition application.\looseness -1

Recent ALPR systems for English VLP captures the employment of promising machine learning models. Such models mainly detects the VLP in the given image (or a video frame) followed by the recognition of the text information on the plate.
A variety of models are stemmed from the need for improving the classification accuracy, robustness to environmental artifacts, and computational efficiency. 
For example, the Convolutional Neural Network (CNN) based bounding box detectors were developed with regression algorithms~\cite{bulan2017segmentation, Xueccv2018, laroca2018, wang2019, silva2021}, manual annotation~\cite{Zhuang_2018_ECCV}, and transfer learning~\cite{c12}. 
Other development with  recurrent neural network (RNN) based architectures include the  Bidirectional Long short-term memory (BiLSTM) based models~\cite{zou2020robust}, and real\hyp time object detection algorithm, YOLO (You Only Look Once) and its variants based models~\cite{chen2019automatic, al2019efficient, yolo}. The end-to-end cascaded and unified architectures of the neural networks were also studied for the ALPR systems~\cite{Hsu2013, Montazzolli2017, Li2019, onim2020traffic}.

However, Unlike the English VLP, the effort in developing the ALPR systems for the Bengali VLP is particularly limited.
Despite much interest in developing Bengali handwriting, scripts and character recognition in general, it has not captured the ALPR system yet. 
A few notable developments of the ALPR systems for the Bengali VLP include the feature extraction based on the digital curvelet transform~\cite{majumdar2007bangla}, 
Tesseract OCR~\cite{hasnat2009integrating}, and CNN with  Adam optimizer~\cite{rabby2018bornonet}.
% presented a new 13-layer CNN model called BornoNet for Bengali characters recognition. The model was designed with two sub-layers optimized using that offered testing accuracy of 95\%.\looseness -1

In summary, no Deep Neural Networks (DNN) model can be tracked in the literature that can detect the Bengali VLP and recognize its characters simultaneously.
A few models only focused on the Bengali characters recognition~\cite{ hasnat2009integrating, rabby2018bornonet}, but their performance is unknown for ALPR system. 
Although Tesseract and BornoNet have relatively high character recognition accuracy, they are not suitable for real\hyp time applications like ALPR due to higher processing time.
Unlike the above models, Onim~\etal\cite{onim2020traffic} recently combined YoloV4 for detection of VLP, and Tesseract as OCR engine.
However, the model requires reasonably higher time for number plate detection and character recognition.
% , and thus it was also not suitable for a real\hyp time application. 
% 
% Thus, for a lower detection accuracy, video frames with VLP will fail to extract the VLP region, and for a poor recognition accuracy, the OCR engine will miss the VLP  for character recognition. 
All these mean that the models  developed for either Bengali VLP detection or Bengali characters recognition are generally limited with \textit{low detection and recognition accuracy}, and \textit{high computational complexity} making them unsuitable for a real~\hyp time application.
% Consequently, 

In this paper, we, therefore, report an ALPR system with a new DNN model that we call Bengali License Plate Network (BLPnet) (Sec.~\ref{sec-our ALPR}).
BLPnet is constructed to have three primary phases.
Particularly, the contributions with the proposed three\hyp phase ALPR system can be summarized as follows: 

\begin{itemize}
    \item Building on the NASNet\hyp Mobile backbone architecture, the first phase employs more \textit{dense} and \textit{pulling} layers on the network\hyp head to identify the vehicles more efficiently with a region of interest bounding\hyp box (Sec.~\ref{subsec-boundingbox}). 
    
    \item The second phase includes an InceptionV3 architecture customized with several \textit{dense} and \textit{pulling} layers to detect the VLP in the bounding\hyp box region (Sec.~\ref{subsec-VLPdetect}). 
    % 
    % The process of detecting the VLP region from the bounding\hyp box containing the detected vehicle is the main unique feature of this work that enhances the efficiency of this technology.
    % 
    \item The third phase introduces a new Bengali\hyp OCR engine (Sec.~\ref{subsec-OCR}).
    % in the BLPnet with  OCR customized with
    % The third stage utilizes our very own customized OCR 
    % to recognize Bengali characters. 
    Unlike the existing ALPR systems that cannot adequately tackle the artifacts like motion\hyp blur and non\hyp uniform shadow, the proposed engine  employs de\hyp blurring and Region Scalable Fitting (RSF) based segmentation, which are optionally invoked (\ie, when characters are not recognized) to better tackle the intensity inhomogeneity.
    \item Additionally, detecting vehicle regions (first phase) prior to VLP (second phase), as considered in the proposed ALPR system, would significantly reduce computational cost and false\hyp positives making the system faster and more accurate. 
\end{itemize}

\section{Related ALPR Systems}
Development of several prominent ALPR systems for English VLP can be tracked in the literature so far. 
Getting the VLP information requires the localization and detection of the VLP. In CNN based approach, features are extracted for VLP and then object localization is done with bounding box detectors and regression algorithms~\cite{Xueccv2018, wang2019}.
Silva~\etal~\cite{silva2021} and Laroca~\etal~\cite{laroca2018} used bounding box regression to localize the object of interest with image coordinates using YOLO. These are One-stage detection networks and relatively faster than other detectors. But these approach is computationally inefficient as it requires training \textit{darknet} backbone of over $27M$ parameters. 

Alternatively, VLP can also be semantically segmented using a either CNN or deep segmentation network. 
Bulan~\etal\cite{bulan2017segmentation} proposed a real\hyp time complete CNN model  having higher adaptability to tackle the scene\hyp variations  requiring minimum possible human\hyp intervention.
An image\hyp based classification was proposed with weak\hyp Snow and strong\hyp CNN classifiers. 
The first classifier was initially used for the region\hyp tagging of the possible number\hyp plates, and the strong\hyp CNN refined the classification outputs for the actual number\hyp plates
Zhuang~\etal~\cite{Zhuang_2018_ECCV} proposed such method where they segmented the VLP and the characters in it. This approach also needs a manual annotation of a characters and LP. Besides these segmentation based networks, segmentation free Networks have also been proposed. 
Wang~\etal~\cite{wang2019} proposed a convolutional Recurrent Neural Network (CRNN) and proposed a multi-task license plate detection and recognition (MTLPR) model.
Zou~\etal~\cite{zou2020robust} used Bi-LSTM and directly localized characters without segmentation.\looseness -1

Both end-to-end cascaded and unified architectures have been used for ALPR system. Li~\etal~\cite{Li2019}
% and Onim~\etal~\cite{onim2020traffic}
used end-to-end trainable LP detection model that remains error-prone to the challenging conditions and also depends heavily on architectural design.
On the other hand, cascaded architectures have the flexibility of tuning and testing. Hsu~\etal~\cite{Hsu2013} and Montazzolli~\etal~\cite{Montazzolli2017} proposed cascaded architecture for real time VLP detection and recognition.

Huang~\etal~\cite{c12} introduced Transfer learning into their CNN model to tackle the limited labeled data for a high detection accuracy of VLP of  93\%.
Later, with a modified YOLOv3 model with 10 CNNs, Chen~\etal\cite{chen2019automatic} attempted to detect and recognize the VLP  with an accuracy of about 98\% and  78\%, respectively in a variety of conditions (\ie, rainy backgrounds, darkness and dimness, and varied colors and saturation of photos). 
% The system detects license plates with an accuracy of about 98.22\% and recognizes them with an accuracy of about 78.22\%.
% 
In addressing several other environmental artifacts like
poor\hyp contrast and noisy images resulting from foggy, distortion and dusty conditions,
Al-Shemarry~\etal\cite{al2019efficient} improved the accuracy of VLP detection using a  binary descriptor with contrast enhancement, leaving their proposed system limited to real\hyp time application with poor frame\hyp rate and higher computational cost.
To tackle the motion artifacts (\ie, blurry input images),
Zou~\etal\cite{zou2020robust} recently proposed a Bi-LSTM model with  robust blur\hyp kernel estimation that offered a reasonably lower accuracy of 79.55\%.
Another recent development with object bounding\hyp box detection for ALPR system is YOLOv4 model~\cite{yolo}, which is
a deep convolutional network that can localize VLPs with immense frame rate and high accuracy.\looseness -1 

% a multi-level extended local binary patterns (M-ELBP) descriptor uses a Gaussian filter  and contrast-limited adaptive histogram equalization (CLAHE) enhancement method  to extract different features from VLP images 
% under difficult conditions like Low/High Contrast, Foggy Condition, Distorted Condition,Dusty Condition and 
% 
% the accuracy was 99.10\% with 550 hidden nodes. 

%(Bi-LSTM) 
Besides, on Bengali character detection and recognition, 
Majumdar~\etal~\cite{majumdar2007bangla} proposed a feature extraction based on the digital curvelet transform.
Separate $k$\hyp nearest neighbor classifiers were trained using the curvelet coefficients of an original image and its morphologically changed copies. 
The overall recognition accuracy of 96.8\% was reported 
while trained for twenty popular Bengali fonts and tested for different font sizes.
% ,  overall accuracy  were achieved during experiment. 
% 
Later, Hasnat~\etal~\cite{hasnat2009integrating} attempted to successfully introduce Bengali script in Tesseract OCR. 
% 
% Tesseract is having capacities of getting prepared for the languages which are not upheld right now or exist however various font styles which are not upheld now can be prepared with customized prepared information for the better acknowledgement. 
% 
Tesseract can learn new languages or alphabets with training along with existing languages for character recognition. 
Recently, Rabby~\etal~\cite{rabby2018bornonet} presented a new 13-layer CNN model called BornoNet for Bengali characters recognition. The model was designed with two sub-layers optimized using adam optimizer
Despite having relatively higher accuracy in the models for Bengali characters recognition, their applications to an ALPR system is not yet studied.
In addressing the potential gap of having a promising ALPR system for Bengali VLP, we therefore develop and present real\hyp time End\hyp to\hyp End  DNN based model that we call BLPnet.

\section{A New ALPR System with BLPnet}
\label{sec-our ALPR}
\begin{figure*}[htp]
\centering
\includegraphics[width=0.95\columnwidth]{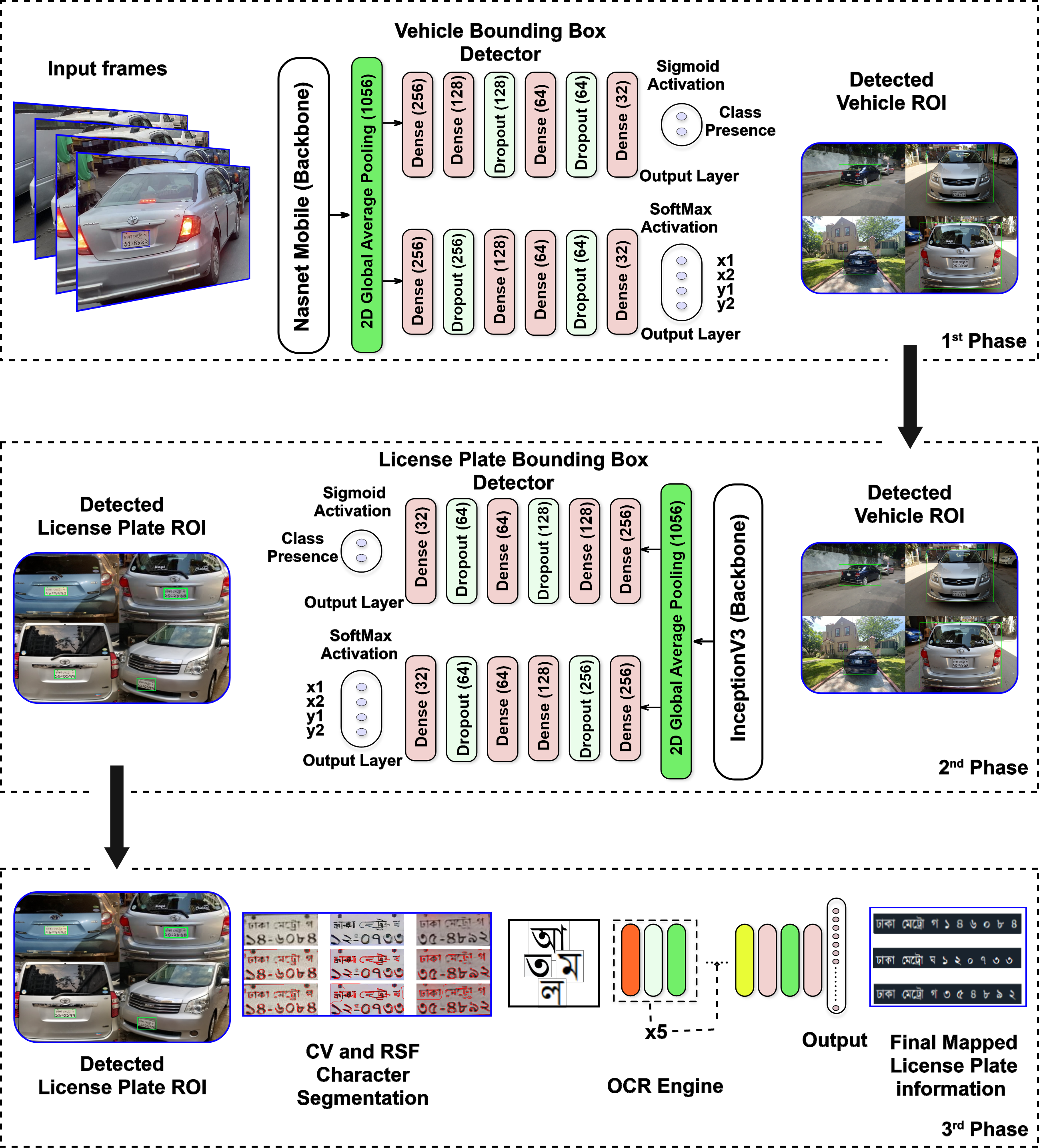}
% \caption{The proposed ALPR system and its three main stages}
% \caption{The proposed ALPR system and its three main stages}
\caption{Key processes of the proposed ALPR system}
\label{Overall block diaram}
\end{figure*}

% Newly added paragraph
% 
This section presents a newly developed BLPnet model for ALPR system.
As illustrated in Fig.  \ref{Overall block diaram}, the model consists of three main phases that are cascaded together to detect vehicle, VLP and recognize the characters in it.
It takes real-time video frames as input consisting of vehicles and its surroundings.
% 
% first stage constructs a vehicle bounding\hyp box detector module. 
In the first phase, the vehicle region is detected and distinguished using a bounding\hyp box 
% s the ROI of vehicle from the frame by drawing a rectangular box enclosing the vehicle. 
as the smallest possible rectangle with vertical and horizontal sides that entirely surround the vehicle captured in a frame. 
% an object is called a bounding\hyp box. If a vehicle is detected in this stage, those 
The frames containing the bounding boxes are fed to the VLP detection model along with the coordinates of the bounding\hyp box. 
Later, the second phase detects and extracts the VLP contour 
followed by the character  localization and recognition in the final phase.
% Finally, the third stage, the extracted image is taken into the character recognition system, where each character is recognised as an object. 
% Thus, producing character localization and recognition output in a single step. 
Consideration of intensity\hyp inhomogeneity resilient segmentation and effective deployment of these cascaded modules would reduce the computational cost and false\hyp positive predictions improving the accuracy and reliability of the system for a real\hyp time application. 
These phases are now discussed below with more technical details.

%  depicts the development and deployment process of the complete ALPR system. 
 
%\begin{figure*}[htb]
%\centering
%\includegraphics[width=0.7\columnwidth]{bbox.png}
% \caption{The proposed ALPR system and its three main stages}
%\caption{Bounding\hyp box detector of the proposed ALPR system}
%\label{fig-BBox}
%\end{figure*}

\subsection{An Extended Vehicle Bounding\hyp Box Detector}
\label{subsec-boundingbox}
\begin{figure*}[htb]
\centering
\includegraphics[width=0.6\columnwidth]{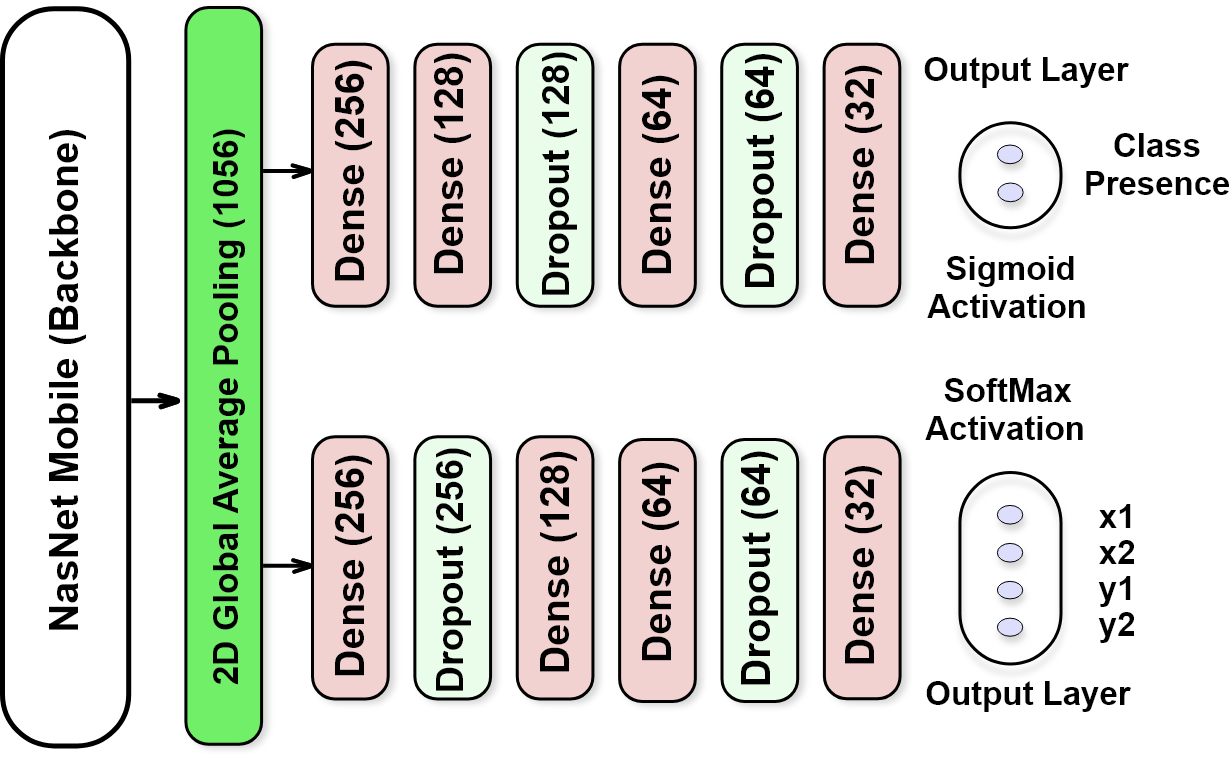}
% \caption{The proposed ALPR system and its three main stages}
\caption{Vehicle Bounding\hyp box detector of the proposed ALPR system}
\label{fig-BBox}
\end{figure*}

%[Here the vehicle detection part will be described with multiple subsections]\\
% 
A NASNet\hyp Mobile is extended and used as the backbone of our DNN followed by six hidden layers to improve the vehicle detection accuracy (see Fig.~\ref{fig-BBox}).
The NASNet\hyp Mobile is more computationally efficient than its counterpart like the ResNet-based architecture that needs higher hardware configuration, making it unsuitable for large\hyp scale deployment~\cite{c13}. 
% our desired result. 
Thus, our model can be implemented on Field-Programmable Gate Array (FPGA) and other embedded devices to balance between hardware resources and processing speed. The TensorFlow (TF) v2.4 framework was used for the overall construction of our CNN model.
The backbone was pre-trained with the Imagenet dataset \cite{imagenet}. 
Three hidden layers added as network head were fully connected layers followed by a dropout layer. A few important processing and considerations are briefly discussed below.\looseness -1 

\paragraph{Dataset Collection}
For the training of vehicle bounding\hyp box detection model, we used the Cars dataset~\cite{stanford} of Stanford AI Lab.
There are 16,185 images in the dataset, representing 196 different car classifications. The data is divided into 8,144 training images and 8,041 testing images, with about a 50-50 ratio between the two classes. Typically, classes are organized by Make, Model, and Year, such as 2012\hyp Tesla\hyp Model\hyp S or 2012\hyp BMW\hyp M3\hyp coupe. Each image in the dataset has also  a  bounding\hyp box output value as an ideal reference that suits our requirement.

\paragraph{Pre-Processing}
We considered a total of  8,041 training images and passed them through two preprocessing stages. At first image augmentation was done to simulate challenging conditions which the real\hyp time traffic video might face. The applied augmentations include: rotate, shift, flip, varying contrast, blur, and salt and pepper noise.
After augmentation, label-encoding and  training-validation split are done. In this case, we made an 80-20 split. After the splitting, the model is trained and tuned.

\paragraph{Transfer Learning (TL)}
A pre\hyp trained model (NASNet-Mobile) as the backbone architecture of the proposed model ensures optimised training.
% according to the recent research trend.
Unlike the isolated learning paradigm, Transfer learning allows extending the pre\hyp learned features to adequately address the new related challenges. 
Particularly, an average pooling layer usually can extract the pretrained models edge detection capability.
From the final layer of shape $(10\times10\times1056)$, an average pooling layer is thus added followed by a group of fully connected dense layers and dropout layers. 
 Finally the output layers identifies the object and generates the bounding box coordinates.
Thus, our model was pre\hyp trained with over fourteen  million images from the ImageNet dataset, which is now capable of categorising images into over 1000 different classes with transfer learning ability.
As a result, it has accumulated a library of rich feature sets for a wide variety of classes.

% \paragraph{Additional Hidden Layers.}
% A global average pooling layer was added on top of the NASNet-Mobile layers, as well as an array of dense and dropout layers. These extra layers generated 637,680 trainable parameters, bringing the total number of trainable parameters to 4,907,396. Finally, the bounding\hyp box coordinates and the class output are provided by these extra layers.

\paragraph{Hyper-parameters and Tuning}
The training hyper-parameters are given in Table~\ref{tab:training_params}. 
% Those parameters are chosen after several trial and error. 
Finally, the parameters with the best accuracy have been chosen after several trials and errors.
The optimizer that suits best for the VLP detection is Stochastic gradient descent(SGD). 
We have also tweaked several parameters of the optimiser 
% to work the best possible way
for our dataset and defined early stopping to reduce the training time. 
Additionally, Reduced Learning Rate functionality has been used to further expedite the training process. 
% The details of which is given in Table~\ref{tab1}.

% \begin{table}[htb]
% \caption{Comparison of the training hyper\hyp parameters}
%   \centering
%     \renewcommand{\arraystretch}{1.05}
%     \setlength{\tabcolsep}{16pt}
%     \resizebox{0.95\linewidth}{!}{%
%     \begin{tabular}{lllll}
% \toprule
%     \textbf{Model} & \multicolumn{2}{l}{\textbf{Our Model}} & \multicolumn{2}{l}{\bf \hl{YOLOv4.0 Model}}\\ \midrule
%     \textbf{Parameters} &	\textbf{Vehicle} &  \textbf{VLP} & \textbf{Vehicle} &  \textbf{VLP}\\
%     \midrule
%     \multirow{2}{*}{\textbf{Backbone}} & \textbf{NASNet-} &	\multirow{2}{*}{\textbf{Inception.V3}} & \multirow{2}{*}{\textbf{Darknet}} & \multirow{2}{*}{\textbf{Darknet}}\\
%     & \textbf{Mobile} &  & & \\\midrule
%     \textbf{classes} & 1 & 1 & 4 & 1\\
%     \textbf{batch} & 32 & 64 & 32 & 64\\
%     \textbf{learning\_rate} & 0.001 & 0.001 & 0.001 & 0.001\\
%     \textbf{epochs} & 1000 & 6000 & 4800 & 5400\\
%     \textbf{image size} & $300\times 300$ & $200\times 200$ & Any & Any\\
%     \textbf{Optimizer} & SGD & Adam & SGD & Adam\\
%     \textbf{Alpha ($\alpha$)} & $10^{-3}$ & $10^{-3}$ & $10^{-3}$ & $10^{-3}$\\
%     \bottomrule
%     % \vline
%     \end{tabular}
%     }
%     % \caption{Caption}
%     \label{tab1}
% \end{table}

\begin{table}[htb]
    \centering
    \caption{Comparison of the training hyper\hyp parameters}
    \setlength{\tabcolsep}{4pt}
    \resizebox{\linewidth}{!}{
    \begin{tabular}{l|cc|cc|c|c|c}
    \toprule
       \bf Model	& \multicolumn{2}{c|}{\bf Ours (BLPnet)}	& \multicolumn{2}{c|}{\bf Onim~\etal~\cite{onim2020traffic}}	& \bf Zou~\etal~\cite{zou2020robust}	& \bf \makecell{Hendry \&\\ Chen~\cite{chen2019automatic}}	& \bf Li~\etal~\cite{LI201814}\\
        \midrule
        \bf Parameters &	Vehicle	& VLP	& Vehicle &	VLP	& VLP	& VLP	& VLP\\
        \midrule
        \bf Backbone	& \makecell[l]{NASNet\\Mobile}	& InceptionV3	& Darknet	& Darknet	& MobileNetV3	& Darknet & - \\
        \midrule
        \bf Classes &	1 &	1 &	4 &	1 &	37 &	60 &	37\\
        \midrule
        \bf Batch size	& 32 &	64 &	32 &	64 &	- &	- &	32\\
        \midrule
        \bf Image Shape &	$300\times300$ &	$200\times200$ &	$224\times224$ &	$128\times128$ &	$152\times56$ &	$256\times256$ &	$24\times24$ \\
        \midrule
        \bf Optimizer &	SGD &	Adam &	Adam &	Adam &	
        SGD &	Adam &	Adam\\
        \midrule
        \bf Epochs &	300 &	300 &	100 &	3500 & - &		600 & - \\
    \bottomrule
    \end{tabular}
    \label{tab:training_params}
    }
\end{table}

\subsection{VLP Detection with Inception-v3}
\label{subsec-VLPdetect}
\begin{figure*}[htb]
\centering
\includegraphics[width=0.6\columnwidth]{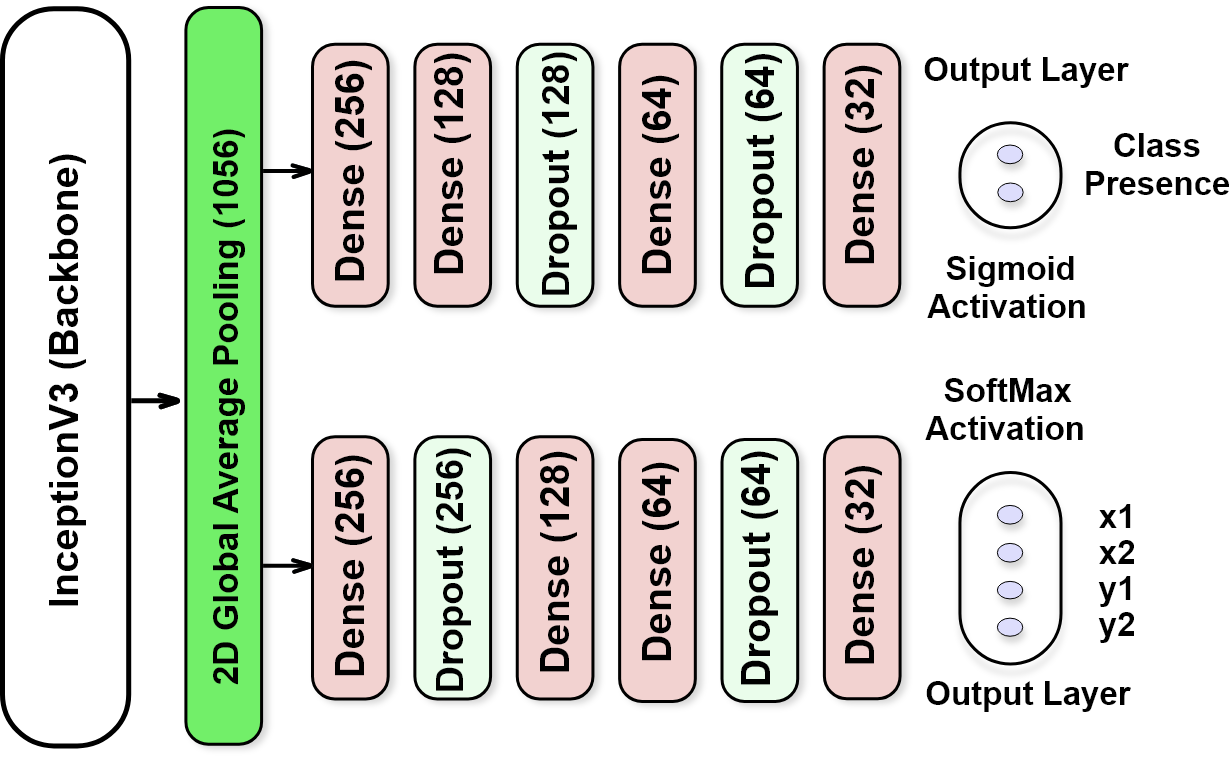}
% \caption{The proposed ALPR system and its three main stages}
\caption{License Plate Bounding\hyp box detector of the proposed ALPR system}
\label{lp-BBox}
\end{figure*}

We have also extended the
% For the purpose of robust object detection, we used
Inception-v3 DNN architecture and employed it as the backbone of the second phase process of BLPnet for VLP detection as illustrated in Fig.~\ref{lp-BBox}.
% The proposed VLP detector detects VLP.
% model for VLP detection. 
Inception-v3 is a CNN designed from the Inception family that transports label\hyp  information to a lower level of the network using label\hyp smoothing, factorised $7\times 7$ convolutions, an auxiliary classifier, and batch\hyp  normalization for layers in the side head. 
Inception architecture being 42 layers deep has 2.5 times lower  computational cost than that of GoogLeNet and can function well even when memory and computational budgets were limited~\cite{szegedy2016rethinking}.

We have used an annotated dataset of 1500 training and 300 validation images to train our model. 
From real\hyp world video footage taken on roadways in Dhaka, Bangladesh, with varying road conditions, including high traffic congestion, the VLP were detected with reasonably higher accuracy (see Sec.~\ref{sec-result}).
% able to acquire considerable results in correctly detecting VLP.
The training began with a number of iterations set at 6000. 
The average loss did not diminish significantly after 2000 iterations,
% and the accuracy line began to show ups and downs at 3000 iterations. 
and after each 1000 cycles, the training was capable of taking weight backup. 
To reduce training duration, we thus used early stopping.
% Additionally, the Reduce Learning Rate capability was employed to improve the training process.

\subsection{A New Bengali\hyp OCR Engine}
\label{subsec-OCR}
This step recognised characters in the extracted VLP from the previous step. 
Similar to~\cite{9003211}, we addressed the character recognition as an object recognition problem. 
By treating characters as objects, the character segmentation and recognition steps are integrated together that consists of a total 5 convolutional 2D layers with 16, 32, 64, 128 and 512 nodes.
Each of them includes a preceding max-pooling layer with a pool size of 2 and a dropout layer with a drop rate of 20\%. 
The dropout layers were used to prevent the model from over-fitting. 
The kernel size of 2 is considered with necessary padding. 
% our padding valid. 
Relu activation function is used for all the backbone layers.
Finally, a global\hyp average pooling layer is included before the output layer followed by two dense layers. 
Two important considerations of this model are the conditional use of de\hyp blurring filter and intensity inhomogeneity invariant segmentation of the character.

% \paragraph{Optional Deblurring Filter}
% 
% One key challenge in VLP recognition is 
Blurring is known to be a key challenge in VLP recognition resulting from wrong focusing or vehicle's motion that requires
% Addressing this problem by 
restoration of the blurred images
for a higher accuracy in character recognition. 
A set of filters like  Wienner filter, Total Variation (TV) deblurring, Haar deblurring and a combination of both is considered  for restoring any blurred image to its estimated original version. 
% 
% Additionally, for the blur problem, wiener filter was also implemented as an optional deblur filter.
% % 
% Particularly, the use of Wiener filter as illustrated in Fig.~\ref{wiener filter for discrete signal} can deblur to minimize the squared error between a desired and estimated random process~\cite{article}.
For an input image,  $Im[n]$, let the output of the filter is  $x[n]$ as in~\eqref{eqn_xn}, where $G(z)$ is the filter function, $N$ is the number of past taps with coefficient $a$, $s[n]$  is the reference signal, and $e[n]$ is the residual error.
The criteria to minimize the mean squared error (MSE) for the coefficient is defined in~\eqref{eq-a}, where $E[.]$ is the expectation operator.
On the other hand, Haar deblurring~\cite{DBLP:journals/corr/abs-1905-01003} uses a deconvolution filter and reduce the MSE as in~\eqref{eq-har}, where  $I(i,j)$ is the expected and $G(i,j)$ is the desired deblurred images. 
% The performance of the above-mentioned algorithms is illustrated in Fig.~\ref{Optional Deblur Filter}.
%  
In our model, we have designed a conditional deblurring process invoked when not enough number of characters are detected in the OCR. 
% This problem is solved with the optional filters iterative. 
An iterative approach is employed for this process with a varying threshold
% The threshold for this iteration was chosen 
based on the Fast-Iterative Shrinkage Thresholding Algorithm (FISTA)~\cite{doi:10.1137/080716542} in the VLP image.\looseness -1

\begin{subequations}
\begin{equation}
\label{eqn_xn}
x[n]=\sum_{i=0}^{N} a_{i} \times \operatorname{Im}[n-1]
\end{equation}
\begin{equation}\label{eq-a}
a_{i}=\operatorname{argmin}\left(E\left[e^{2}[n]\right]\right)\\
\end{equation}
\end{subequations}\vspace{2pt}

\begin{equation}\label{eq-har}
M S E=\frac{1}{m \times n} \sum_{i=0}^{m-1} \sum_{j=0}^{n-1}[I(I, J)-G(i, j)]^{2}
\end{equation}

% \paragraph{Character Segmentation}
% 
Additionally, we have considered the intensity inhomogeinity invariant segmentation of the characters. Thus, we extracted the characters from the VLP with two segmentation models: Chan~\&~Vese (CV) model and Region-Scalable Fitting (RSF) model (see Table~\ref{Table for Performance Analysis of OCR} in Sec.~\ref{sec-result}). As most of the images for VLP have very high contrast, CV model performed well in approximating the intensities of the foreground or object and background. 
In case of intensity inhomogeneity, CV model failed to segment the characters, where RSF model was used.

\begin{figure*}
    \centering
    \includegraphics[width=0.8\linewidth]{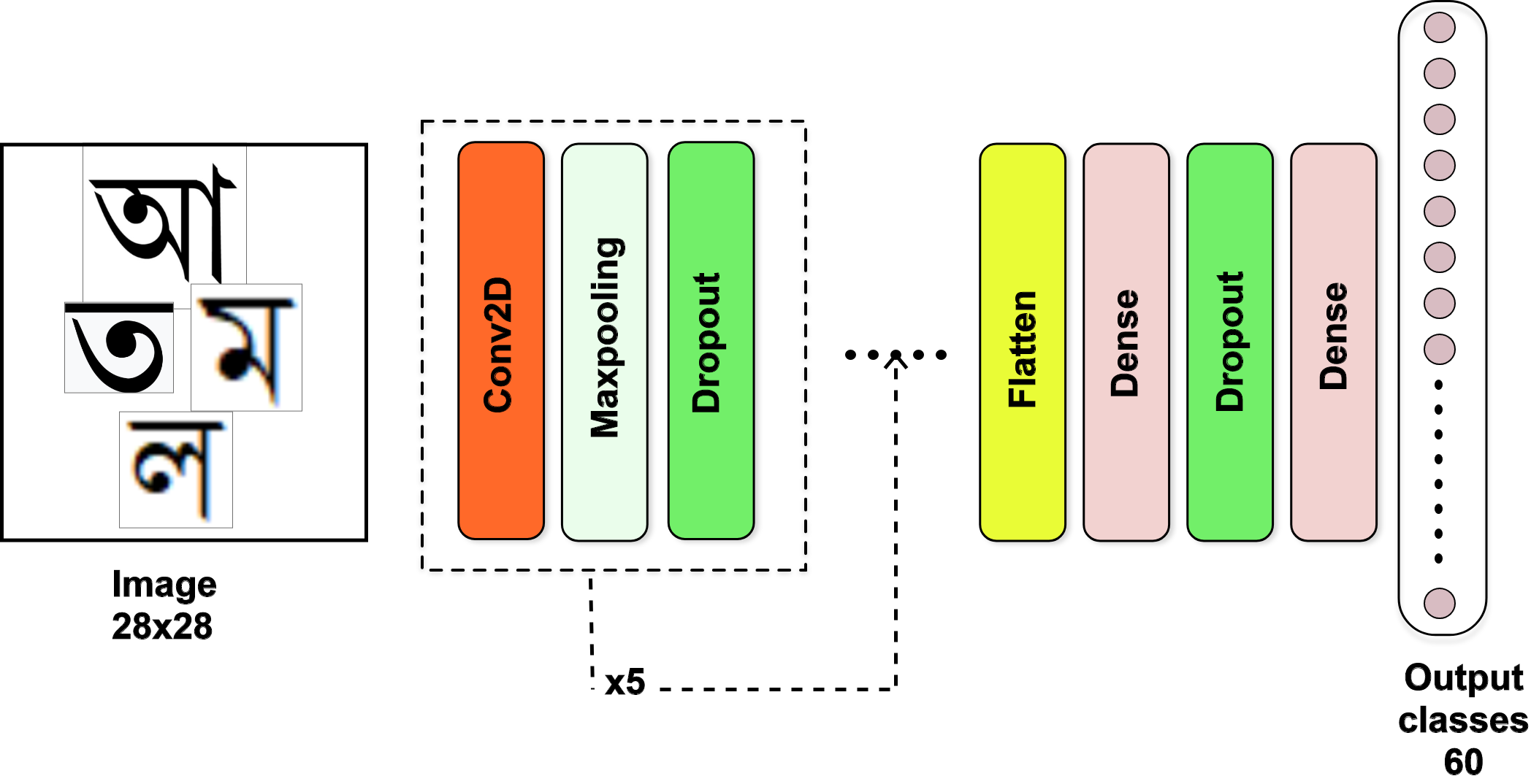}
    \caption{The newly developed OCR engine for BLPnet}\vspace{-3pt}
    \label{OCR} 
\end{figure*}

The complete architecture is illustrated in Fig.~\ref{OCR}.  
The output layer contains 60 classes to capture the total number of Bengali alphabets and digits. 
The output layer uses \textit{Softmax} as an activation function. 
We have also used CMATERdb version 3.1.2 for training the alphabets for the digits. A total alphabet\hyp images of  15000 and digits of 500 have been split as follows: 9900 images for training,   2600 images for validation, and 3000 images for testing.

\begin{figure}[thb]
\centerline{\includegraphics[width=0.55\linewidth]{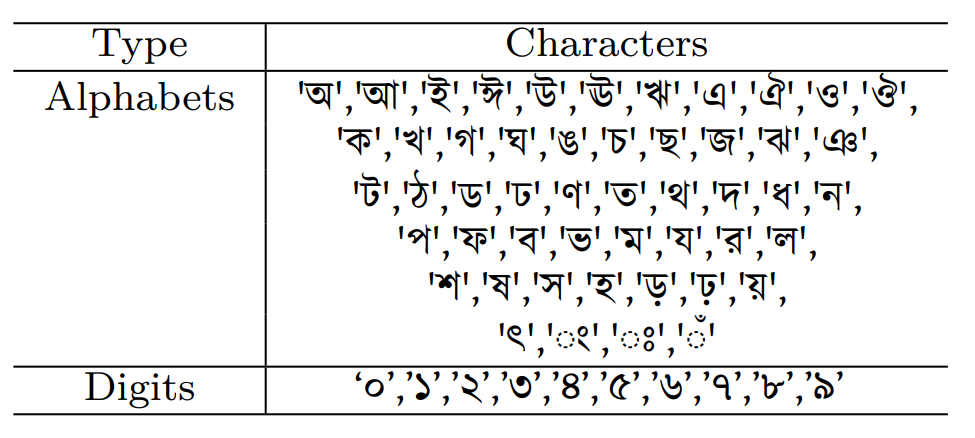}}
\caption{Bengali character classes for the OCR training}\vspace{-3pt}
\label{Character Class}
\end{figure}

The images are then augmented so that the model can identify images on non-ideal conditions. The following augmentations are considered: (\textit{i})~10\% shift in left and 10\% shift in right; (\textit{ii})~10\% shift in vertical and 10\% shift in horizontal; (\textit{iii})~7.5$\degree$ rotation in left and 7.5$\degree$ in right; (\textit{iv})~zoom range 20\%; and, (\textit{v})~Shear range 20\%.
Besides, each character illustrated in Fig.~\ref{Character Class} is resized to $16\times 16$ after segmentation before passing through the network. 
As a result,  this network can work with variable size characters.
% contains all the characters and digits.

% \begin{figure}[thb]
% \centerline{\includegraphics[width=0.55\linewidth, height=0.7in]{wiener_filter.png}}
% \caption{Wiener filter for discrete signal}
% \label{wiener filter for discrete signal}
% \end{figure}

% \begin{figure}[thb]
% \centerline{\includegraphics[width=0.83\linewidth,height= 2.5in]{deblurr.png}}
% \caption{Performance of the deblurring filter used in our OCR engine}
% \label{Optional Deblur Filter}
% \end{figure}

\section{Results and Analysis}
\label{sec-result}
We now present the performance of the proposed BLPnet model for the ALPR system, and compare its performance with some related models:  %Rabby~\etal~\cite{rabby2018bornonet},
Li~\etal~\cite{LI201814}, Hendry \& Chen~\cite{chen2019automatic}, Zou~\etal~\cite{zou2020robust}, and Onim~\etal~\cite{onim2020traffic}. 
This section starts with the  analysis of total learnable parameters of the model followed by the
 results discussed in three main parts, namely, for vehicle detection, VLP detection and finally, for the OCR and word\hyp mapping.

\subsection{Total Learnable Parameters}
A dense layer with \textit{m} input nodes and \textit{n} output nodes will have a total of $(n+1)\times m$ learnable parameters.
The pooling layers and dropout layers does not learn anything. For convolutional layers with \textit{p} feature maps in input and \textit{q} feature maps as output having a filter size of $i\times j$ will have a total learnable parameters of $(i\times j\times p + 1)\times q$.
 The distribution of trainable parameters for bounding box detector across the network is noted in Table.~\ref{tab:bbox_param}. 
 The total number of trainable parameters is calculated as shown in Equation.~\eqref{param}. 
 Similarly, Table.~\ref{tab:ocr_params} shows the trainable parameters for our OCR engine. Here \textit{Layers} denotes the name and operation of added layers, \textit{Shape} denotes the input shape of tensors for that particular layer and finally the trainable parameter for that layer.

 \begin{table}[!th]
     \centering
     \caption{Trainable parameters for bounding box detector}
     \renewcommand{\arraystretch}{1.05}
	\setlength{\tabcolsep}{10pt}
	\resizebox{0.7\linewidth}{!}{
     \begin{tabular}{llc}
\toprule
\bf Layers &\bf Shape &\bf \makecell[c]{Trainable\\ Parameters} \\
\midrule
Global\_AveragePooling2D & (None, 1056) & 0 \\
Dense & (None, 256) & 270592 \\
Dense & (None, 256) & 270592 \\
Dense & (None, 128) & 32896 \\
Dropout & (None, 256) & 0 \\
Dropout & (None, 128) & 0 \\
Dense & (None, 128) & 32896 \\
Dense & (None, 64) & 8256 \\
Dense & (None, 64) & 8256 \\
Dropout & (None, 64) & 0 \\
Dropout & (None, 64) & 0 \\
Dense & (None, 32) & 2080 \\
Dense & (None, 32) & 2080 \\
Dense & (None, 2) & 66 \\
Dense & (None, 4) & 132 \\
\midrule
\bf Total & & 627k\\
\bottomrule
\end{tabular}
}
     \label{tab:bbox_param}
 \end{table}

\begin{equation}
\begin{aligned}
    N_{detector} &=2\times((1056+1) \times 256+(256+1) \times 128+(128+1) \times 64 \\
    & +(64+1) \times 32+(32+1) \times 2) \\
    &=627k
\end{aligned}
\label{param}
\end{equation}

\begin{table}[!t]
    \centering
    \caption{Trainable parameters for OCR engine}
\renewcommand{\arraystretch}{1.05}
	\setlength{\tabcolsep}{8pt}
	\resizebox{0.6\linewidth}{!}{
\begin{tabular}{llc}
\toprule  
\bf Layers &  \bf Shape & \bf \makecell[c]{Trainable\\ Parameters}  \\
\midrule
Conv2D  & (None, 63, 63, 16) & 80 \\
MaxPooling2D & (None, 31, 31, 16) & 0 \\
Dropout & (None, 31, 31, 16) & 0 \\
Conv2D  & (None, 30, 30, 32) & 2080 \\
MaxPooling2D & (None, 15, 15, 32) & 0 \\
Dropout & (None, 15, 15, 32) & 0 \\
Conv2D  & (None, 14, 14, 64) & 8256 \\
MaxPooling2D & (None, 7, 7, 64) & 0 \\
Dropout & (None, 7, 7,64)  & 0 \\
Conv2D  & (None, 6, 6, 128) & 32896 \\
MaxPooling 2D & (None, 3, 3, 128)  & 0 \\
Dropout  & (None, 3, 3, 128)  & 0 \\
Conv2D  & (None, 2, 2, 256)  & 131328 \\
MaxPooling 2D  & (None, 1, 1, 256)  & 0 \\
Dropout  & (None, 1, 1, 256)  & 0 \\
Flatten  & (None, 256) & 0 \\
Dense & (None, 256)  & 65792 \\
Dense & (None, 512)  & 131584 \\
Dropout & (None, 512) & 0 \\
Dense  & (None, 60) & 25650 \\
\midrule
\bf Total &  & 397K\\
\bottomrule
\end{tabular}
}
    \label{tab:ocr_params}
\end{table}

% As most of the images for VLP have a very high contrast the segmentation was quite easy for any moderate algorithm. Chan~\&~Vese (CV)~\cite{902291} proposed an active contour model (CV) by assigning two constants to approximate the intensities of the foreground or object and background. Li~\etal~\cite{DING2017224} developed a region-scalable fitting model (RSF) for segmentation that solves the intensity in-homogeneity problem. 
% Table~\ref{Table for Performance Analysis of OCR} demonstrates the effectiveness of these two methods measured against time and accuracy.

\subsection{Vehicle Detection}
% After several trials and errors, the best possible hyper-parameters were recorded. 
We trained the vehicle bounding\hyp box model for vehicle detection for 1000 epochs with early stopping employed to stop the model training if the accuracy improvement is negligible. 
The hardware specification to train the model is: Tesla K80 2496 CUDA cores GPU with 12~GB GDDR5 VRAM, Processor @2.3Ghz (4 core, 8 threads), and RAM of 48~GB.

% and the learning rate is reduced further to get the best possible result.
\begin{figure}[!th]
\centerline{\includegraphics[width=.8\linewidth]{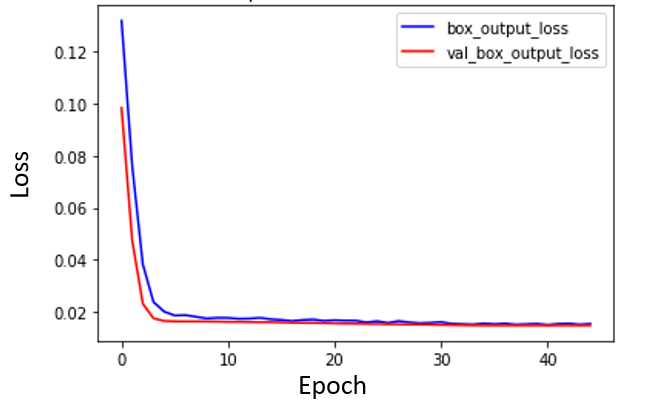}}
\caption{Loss vs Epoch curve for the first 50 epochs}
\label{Loss vs Epoch curve}
\end{figure}

\begin{figure*}[htbp]
\centering
    \subfloat[Detected vehicles (marked with green bounding\hyp boxes) ]{%
    \includegraphics[width=0.9\columnwidth, height = 3 in]{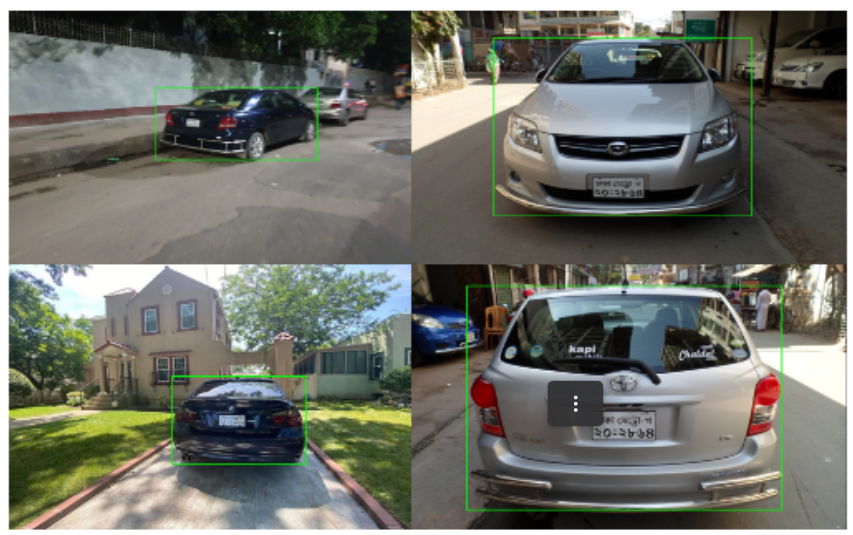}\label{Bounding box output}}
    \\
    \subfloat[VLP detection (marked with blue\hyp colour bounding box)]{%
    \includegraphics[width=0.9\columnwidth, height = 3in]{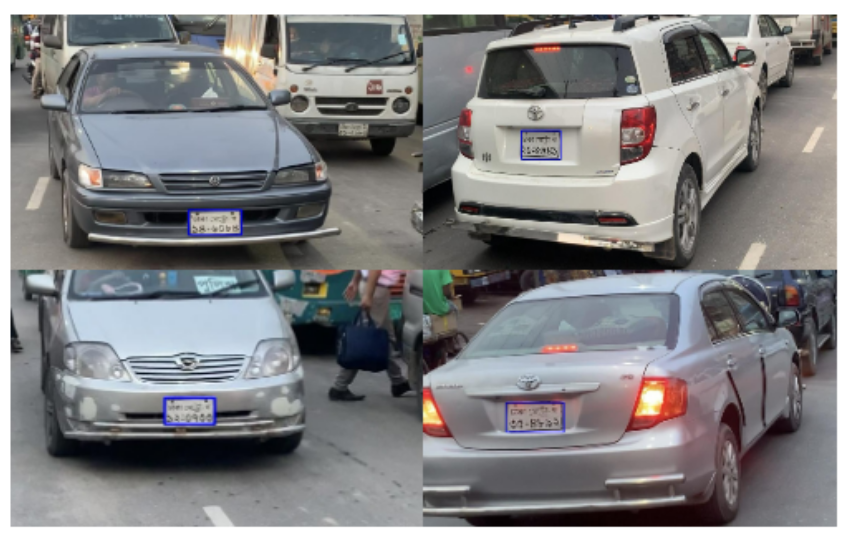}\label{cars}}
\caption{Example of detected vehicles and VLP from real\hyp time captured video footage with different orientations of the vehicles}
% \caption{Example of VLP detection (marked with blue\hyp colour bounding box) for different orientations of the vehicles}
\end{figure*}

% \begin{figure}[t]
% \centerline{\includegraphics[width = 0.55\textwidth]{confusion_matrix.png}}
% \caption{Confusion matrix for the alphabets}
% \label{conf_matrix}
% \end{figure}

We chose the criteria for our evaluation matrices of bounding\hyp box coordinates to be MSE and accuracy.
From the loss vs epoch curve in Fig.~\ref{Loss vs Epoch curve}, we observe that the model can quickly converge to a lower loss value. After a while, the improvement gets very slow.
The best training loss and MSE of bounding\hyp box were 0.0130 and 0.0155, respectively and the validation loss and MSE were 0.0148 and 0.0152, respectively. 
Overall training parameters with their respective epochs is presented in Table~\ref{table of vb training}. 
We observe in this table that after 300 epochs, the rate of change in loss and accuracy is becoming stable.
Compared to the YOLO model, the hyper-parameters were kept similar as shown earlier in Table~\ref{tab:training_params}.
All these observations suggest that the our proposed model would effectively detect the vehicles in the input video clips. The final predicted output of the model applied in random vehicles is shown in Fig.~\ref{Bounding box output} with successful demarcation of green boxes.

\begin{table*}[!th]
\caption{Performance of vehicle detection with bounding\hyp box}
\centering
\renewcommand{\arraystretch}{1.05}
	\setlength{\tabcolsep}{3pt}
	\resizebox{\linewidth}{!}{
\begin{tabular}{cccccccccc}
\toprule
\multirow{3}{*}{\textbf{Epochs}} & 
\multirow{3}{*}{\textbf{Loss}} & \textbf{Box} & \textbf{Class} & \textbf{Box} & \textbf{Class} & \multirow{3}{*}{\makecell{\bf Validation \\ \bf Loss}} & \makecell{\bf Validation\\ \bf box} & \makecell{\bf Validation\\\bf  class} &  \makecell{\bf Validation\\ \bf box} \\
 &  & \textbf{Output} & \textbf{Output} & \textbf{Output} & \textbf{Output} &  & \textbf{Output} & \textbf{Output} & \textbf{Output} \\
  &  & \textbf{Loss} & \textbf{Loss} & \textbf{MSE} & \textbf{Accuracy} &  & \textbf{Loss} & \textbf{Loss} & \textbf{MSE} \\
    \midrule

50&
0.0056 &
0.0176 &
0.0059 &
0.0176 &
75.59 &
0.0057 &
0.0175 &
0.0057 &
0.0175 \\

100&
0.0056 &
0.0171 &
0.0058 &
0.0171&
85.89 &
0.0056 &
0.0140 &
0.0057 &
0.0140 \\

150&
0.0056 &
0.0169 &
0.0058 &
0.0169 &
89.19 &
0.0055 &
0.0138 &
0.0056 &
0.0138 \\

200&
0.00569 &
0.0173 &
0.0057 &
0.0173 &
90.55 &
0.0055 &
0.0175 &
0.0055 &
0.0175 \\

250&
0.00569 &
0.0176 &
0.0056 &
0.0176 &
94.74 &
0.0054 &
0.0175 &
0.0055 &
0.0175 \\

300&
0.00569 &
0.0176 &
0.0053 &
0.0176 &
96.99 &
0.0055 &
0.0175 &
0.0054 &
0.0175 \\

350&
0.00569 &
0.0176 &
0.0052 &
0.0176 &
97.02 &
0.0055 &
0.0175 &
0.0054 &
0.0175 \\

400&

0.00582 &
0.0130 &
0.0050 &
0.0155 &
97.05 &
0.0056 &
0.0148 &
0.0058 &
0.0152 \\
\bottomrule
% \vline
\end{tabular}
}
% \caption{Caption}
\label{table of vb training}
\end{table*}

% \begin{figure}[t!]  %subfigg
%     \centering
  
%     \hfill
%     \subfloat[flipped\label{fig:ex1gt}]{%
%         \includegraphics[width=0.49\linewidth]{image1-flip.jpg}}
%     \hfill
%     \subfloat[rotated\label{fig:ex1resFast}]{%
%         \includegraphics[width=0.49\linewidth]{image2-rotation.jpg}}
%     \hfill
%     \subfloat[contrast]{%
%         \includegraphics[width=0.49\linewidth]{image3-contrust_change.jpg}}
%     \hfill
%     \subfloat[blur]{%
%         \includegraphics[width=0.49\linewidth]{image4-blur.jpg}}
%     \hfill
%     \subfloat[Salt and Pepper Noise]{%
%         \includegraphics[width=0.49\linewidth]{image5-salt-paper.jpg}}
%     \hfill
%     \subfloat[left shift]{%
%         \includegraphics[width=0.49\linewidth]{image7-left_shifting.jpg}}
%     \hfill
%     \subfloat[right shift]{%
%         \includegraphics[width=0.49\linewidth]{image6-right_shifting.jpg}}
    
%     \caption{all types of image augmentation listed herel}
%     \label{image augmentation}
% \end{figure}

% \begin{figure}[thb]
% \centerline{\includegraphics[width=0.83\linewidth,height= 2.5in]{deblurr.png}}
% \caption{Performance of the deblurring filter used in our OCR engine}
% \label{Optional Deblur Filter }
% \end{figure}

\subsection{VLP Detection}

Our model was evaluated using both real\hyp time and pre-recorded video clips. Table \ref{tab:training_params} shows the hyper\hyp parameters used to train the network. Our model did not over\hyp fit due to the use of adequate \textit{dropout} and \textit{pooling} layers, and thus, it took little time to train and converge to an optimum detection level. During the evaluation, the performance of the model was monitored in real\hyp time. 

Our algorithm successfully detected VLP as a few examples are illustrated in Fig.~\ref{cars}, where VLP is marked with blue bounding box. 
A VLP was detected from varying orientation of the detected vehicle and at that time, 17 frames per second processing speed was maintained on average.
Such accurate detection was observed for all the testing video clips with resolution $1920\times  1080$ and  frames per second (\textit{fps}) is of 15 and 20.\looseness -1 

Similar to the vehicle detection with  bounding box (Sec.~\ref{subsec-boundingbox}), the VLP detection training loss and MSE of bounding\hyp box were 0.013 and 0.016, respectively and the validation loss and MSE were 0.015 and 0.015, respectively.
No false\hyp positive detection was recorded during the process which justifies the effectiveness of previous phase.  \looseness-1

\subsection{Character Recognition}
\begin{figure}[!t]  %subfig
    \centering
 \vspace{-0.3cm}
    \includegraphics[width=0.985\linewidth]{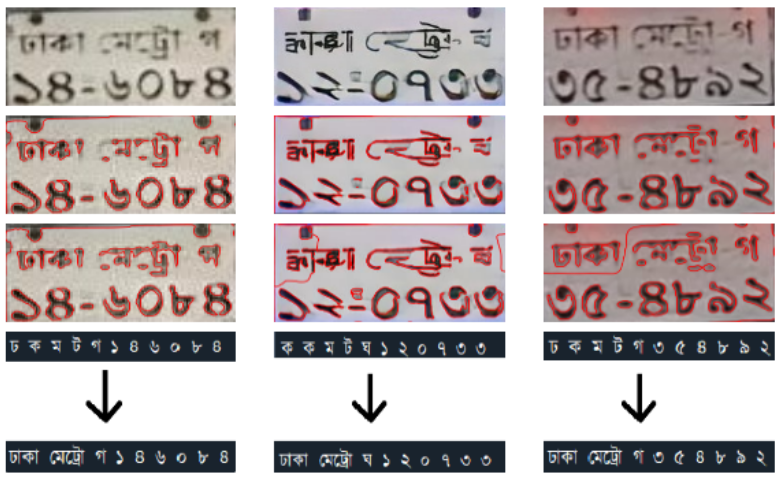}
    \caption{Example of input and outputs of the newly developed OCR engine (From \textit{top}, input images in \textit{first row}, CV\hyp segmented images in \textit{second row},  LSF\hyp segmented images in \textit{third row}, and OCRs and word\hyp mapping in \textit{fourth row})}
    \label{Example results OCR} 
\end{figure}

% newly added
The overall procedure can be broken down into three steps. First the preprocessing step, where the image is filtered based on the sharpness. 
% Fig. \ref{Optional Deblur Filter} depicts the deblur filter's result. 
Later in character segmentation, the fast active contour model segments the characters as objects. Finally, the character recognition followed by a word mapping. Here, for each  group of key characters recognised, a predefined word is mapped for them. Our model's generated character output is shown in Fig.~\ref{Example results OCR}.

\begin{table}[!t]
\caption{OCR performance for the CV and RSF based segmented characters}
\centering
\renewcommand{\arraystretch}{1}
	\setlength{\tabcolsep}{12pt}
	\resizebox{0.95\columnwidth}{!}{
    \begin{tabular}{ccccc}
    % \vline
\toprule
\textbf{Segmentation} & \textbf{No of} &
\textbf{Accuracy} &
\textbf{Time taken} &
\textbf{Time taken}\\
\textbf{Model} & \textbf{characters} &
\textbf{of OCR} &
\textbf{for OCR} &
\textbf{for Tesseract}\\
& \textbf{extracted} & \textbf{(in \%)} & \textbf{(in seconds)} & \textbf{(in seconds)}
\\
    \midrule
\textbf{CV model} & 4 & 90 & 0.302 & 0.402\\
& 5 & 83 & 0.395 & 0.548\\
& 6 & 81 & 0.432 & 0.701\\
& 8 & 80 & 0.502 & 0.705\\ \midrule
\textbf{RSF model} & 4 & 95 & 0.256 & 0.402\\ 
& 5 & 93 & 0.312 & 0.548\\
& 6 & 90 & 0.333 & 0.701\\
& 8 & 89 & 0.398 & 0.705\\\bottomrule
    % \vline
    \end{tabular}
    }
    % \caption{Caption}
    \label{Table for Performance Analysis of OCR}
\end{table}

\begin{table}[!t]
    \caption{Average testing accuracy in characters recognition applications}
    \centering
    \renewcommand{\arraystretch}{1.2}
    \setlength{\tabcolsep}{10pt}
    \resizebox{0.95\linewidth}{!}{%
    \centering
    \begin{tabular}{lll}
        \toprule \multicolumn{1}{l}{ \bf Model } & \bf Accuracy & \bf Recognising object \\
        \midrule
        % Rabby~\etal~\cite{rabby2018bornonet} & $95.0 \%$ & Handwritten characters (Bengali)\\
        Hendry \& Chen~\cite{chen2019automatic} & $78.2\%$ &  VLP \& characters (English)\\
        Zou~\etal~\cite{zou2020robust} & $79.5\%$ & Blurry text recognition (English)\\
        Onim~\etal~\cite{onim2020traffic} & $90.51\%$ & VLP (Bengali)\\
        % Majumder~\etal~\cite{majumdar2007bangla} & $96.8 \%$ \\
        % Li~\etal~\cite{LI201814} & $????$ &  VLP \& characters (English) \\
        \textbf{BLPnet (proposed)} & $\mathbf{95.0\%}$ & Real\hyp time VLP \& characters (Bengali)\\
        \bottomrule
    \end{tabular}
    }
    \label{tab_comp}
\end{table}

\begin{table}[!t]
    \centering
    \caption{Comparison of features of the ALPR systems}
    \setlength{\tabcolsep}{8pt}
	\resizebox{\linewidth}{!}{
    \begin{tabular}{lp{3cm}cccp{3cm}}
\toprule          
\bf Model  &
\bf  \makecell[c]{VLP\\False--Positive\\Detection} & 
\bf \makecell{Character \\rotational \\ invariant?} &  
\makecell{\bf Trainable \\\bf parameters\\ \bf ({\it \textbf{Million}})} & \bf \makecell{Processing \\time\\ (\textit{s})}&
\bf Hardware\\
\midrule
\makecell[l]{Hendry \&\\ chen~\cite{chen2019automatic}} & Unknown & no & 8.8 & 0.8 -- 1.0 & \makecell[l]{Core i7\\ Nvidia GTX\\ 970 4GB GPU}\\
\midrule
Li~\etal~\cite{LI201814} & 
\makecell[l]{Heuristically \\minimized \\with CNN} & no & 1.0 &
2.0 -- 3.0 & \makecell[l]{Core i5\\ Nvidia Tesla\\ k40c 4GB GPU}\\
\midrule
Zou~\etal~\cite{zou2020robust} & Unknown & yes & 1.9 & Unknown & \makecell[l]{Core i5\\ Nvidia Titan\\ 12GB GPU}\\
\midrule
Onim~\etal~\cite{onim2020traffic} & \makecell[l]{Eliminated using\\ 2\hyp phase detection} & no & 27& 0.7 -- 1.0 &\makecell[l]{Core i5\\ Nvidia Tesla\\ T4 6GB GPU} \\
\midrule
\bf \makecell[l]{BLPnet \\(Ours) }& \bf \makecell[l]{Eliminated using\\ 2\hyp phase detection}  & \bf yes & \bf  0.97 &
\bf 0.32 -- 0.52  & \bf  \makecell[l]{Core i5\\ Nvidia Tesla \\k80 6GB GPU}\\
\bottomrule
\end{tabular}
    \label{tab:major_comp}
    }
\end{table} 
 
% \begin{table}[!ht]
%     \caption{Average testing accuracy in characters recognition applications}
%     \centering
%     \renewcommand{\arraystretch}{1.07}
%     \setlength{\tabcolsep}{10pt}
%     \resizebox{0.95\linewidth}{!}{%
%     \centering
%     \begin{tabular}{llll}
%         \toprule \multicolumn{1}{l}{ \bf Model } & \bf Accuracy & \bf Dataset &\bf Recognising object \\
%         \midrule
%         % Rabby~\etal~\cite{rabby2018bornonet} & $95.0 \%$ & Handwritten characters (Bengali)\\
%         Hendry \& Chen~\cite{chen2019automatic} & $78.2\%$ & AOLP &VLP \& characters (English)\\
%         Zou~\etal~\cite{zou2020robust} & $79.5\%$ & CLPD&  Blurry text recognition (English)\\
%         Onim~\etal~\cite{onim2020traffic} & $90.51$ &CARS & VLP\\
%          Li~\etal~\cite{LI201814} & $????$ & AOLP &VLP \& characters (English) \\
%         \textbf{BLPnet (proposed)} & $\mathbf{95.0\%}$ & \bf CMATERdb & Real\hyp time VLP \& characters (Bengali)\\
%         \bottomrule
%     \end{tabular}
%     }
%     \label{tab_comp}
% \end{table}

Our proposed OCR engine demonstrated reasonably better accuracy  as it treated the characters as objects. The accuracy varies from 80\% to 95\% based on the number of characters extracted. The time it takes to extract a character varies between 0.302 to 0.502 seconds. The model is weak against some vowel pairs. This weakness was tackled successfully with word mapping (see Fig.~\ref{Example results OCR}). Detailed performance analysis is given in Table~\ref{Table for Performance Analysis of OCR}. The accuracy decreases with the increase in number of characters. 
A comparison with related OCR engines is shown in Table~\ref{tab_comp}.
The other performance of the proposed model is also compared  with the other prominent models in Table~\ref{tab:major_comp}.
Considering the nullified False\hyp Positives, lower computational complexity, reduced trainable parameters, and reasonably higher accuracy, the proposed model has demonstrated a higher potential for a real\hyp time ALPR application.

 \section{Conclusion}
 The development of an ALPR system is a timely requirement for modern transportation services. 
 However, despite being the sixth largest population  around the world~\cite{ ethnologue2021}, no significant progress can be tracked in the Bengali language countries or states for the ALPR system addressing their inadequate road\hyp safety measures and poor traffic management. 
% Besides,  research  endeavors continue  to  reduce  computational  complexities  with  higher  detection accuracy and robustness to environmental artifacts. 
% 
%  One such potential group of the developing regions is the Bengali speaking  countries and states that still requires a promising ALPR system for the Bengali VLP application.
 To this end, we have presented a  computationally efficient and reasonably accurate BLPnet model for a new ALPR system.
 The model is designed to efficiently and correctly return the VLP number by considering a three\hyp phase top\hyp to\hyp bottom approach. 
 This means, the model starts from detecting the vehicle first, and then any possible VLP followed by the recognition of characters in the VLP.
This consideration is observed to be more  effective with lower-computational time and accuracy than the prominent ALPR systems.\looseness -1 

Moreover,  the model generates the actual license number of the vehicle from the recognized characters using our simple, yet effective mapping algorithm with a set of predefined cases of registration area\hyp codes.
% , the proposed model can also successfully map the actual vehicle registration numbers on the VLP . 
The model also performed well without compromising accuracy to tackle challenging conditions that cause rotated,  blurry or noisy frames at the input. 
The preliminary results with the reasonably faster and accurate performance of the model suggest that it would be promising for the real\hyp time ALPR application with a continuing development in future.\looseness -1

 \bibliographystyle{elsarticle-num}

\end{document}